# A Complete Calculus for Possibilistic Logic Programming with Fuzzy Propositional Variables

**Teresa Alsinet**
Computer Science Dept.
Universitat de Lleida (UdL)
25001 Lleida, Spain
tracy@eup.udl.es

**Lluís Godo**
AI Research Inst. (IIIA)
CSIC
08193 Bellaterra, Spain
godo@iiia.csic.es

## Abstract

In this paper we present a propositional logic programming language for reasoning under possibilistic uncertainty and representing vague knowledge. Formulas are represented by pairs $(\varphi, \alpha)$, where $\varphi$ is a many-valued proposition and $\alpha \in [0, 1]$ is a lower bound on the belief on $\varphi$ in terms of necessity measures. Belief states are modeled by possibility distributions on the set of all many-valued interpretations. In this framework, (i) we define a syntax and a semantics of the general underlying uncertainty logic; (ii) we provide a modus ponens-style calculus for a sublanguage of Horn-rules and we prove that it is complete for determining the maximum degree of possibilistic belief with which a fuzzy propositional variable can be entailed from a set of formulas; and finally, (iii) we show how the computation of a partial matching between fuzzy propositional variables, in terms of necessity measures for fuzzy sets, can be included in our logic programming system.

## 1 INTRODUCTION

Logic programming languages have been applied to a wide range of areas such as Artificial Intelligence or Deductive Databases. Among these languages, Prolog is the most representative but it is not powerful enough for reasoning and representing knowledge in situations where there is vague, incomplete or imprecise information. To overcome this problem, new logic programming languages have been developed. They are based on a variety of non-standard logics such as multiple-valued logics [Ishizuka and Kanai, 1985; Mukaidono et al., 1989; Li and Liu, 1990; Alsinet and Manyà, 1996; Vojtas, 1998], possibilistic logic [Dubois et al., 1991], probabilistic logic [Heinsohn, 1994; Lukasiewicz, 1998], evidential logic [Baldwin, 1987; Baldwin et al., 1995] or fuzzy operator logic [Weigert et al., 1993]. Depending on the underlying logic some systems are more suitable for dealing with vague knowledge, while others are more appropriate for reasoning under incomplete or imprecise knowledge. Although all of the fuzzy extensions of logic programming implement proof procedures for fuzzy reasoning, only some of them allow to represent ill-known information in the language.

Our first objective in this paper is to define a propositional logic programming language for reasoning under possibilistic uncertainty and representing vague knowledge. We represent formulas by pairs $(\varphi, \alpha)$, $\varphi$ being a many-valued proposition built on fuzzy propositional variables and $\alpha \in [0, 1]$ being a lower bound on the belief on $\varphi$ in terms of necessity measures.

On the one hand, fuzzy propositional variables provide us a suitable representation model in situations where there is vague, incomplete or imprecise information about the real world. For instance, the fuzzy statement "Peter is $about\_35$ years old" can be nicely represented by the fuzzy proposition $Peter\_is\_about_{35}$ defined over the finite domain years_old. In the case $about\_35$ denotes a crisp interval of ages, the above proposition can be interpreted as "$\exists x \in about_{35}$ such that Peter is $x$ years old". In the case $about\_35$ denotes a fuzzy interval with a membership function $\mu_{about\_35}$, the above proposition can be interpreted in possibilistic terms as "$\exists x \in [\mu_{about\_35}]_\alpha$ such that Peter is $x$ years old is certain with a necessity of at least $1 - \alpha$" for each $\alpha \in [0, 1]$, where $[\mu_{about\_35}]_\alpha$ denotes the $\alpha$-cut of $\mu_{about\_35}$. So, fuzzy propositions can be seen as (flexible) restrictions on an existential quantifier (see [Dubois et al., 1998]).

On the other hand, since we want to deal with fuzzy propositional variables in the language, the truth evaluation of formulas cannot be Boolean but many-valued, and thus, our possibilistic logic programming language should be based on a many-valued logic as in [Alsinet et al., 1999]. Moreover, like in classical propositional logic programming systems, the language should enable us to define an efficient proof method based on a complete calculus for determining the maximum degree of possibilistic belief with which a fuzzy propositional variable can be entailed from a set of formulas. To this end, first we define a general possibilistic logic based on the propositional

Gödel fuzzy logic and then, we focus our attention on the possibilistic language that results from considering the Horn-rule sublogic of Gödel fuzzy logic.

The reason for choosing Gödel logic as the underlying many-valued logic where to model fuzziness is two-fold: first, truth-functions of Gödel logic are purely ordinal, that is, they are definable just from the ordering of the truth scale (see next section), no further algebraic operations are required, and thus the use of this logic is in accordance with the simplest understanding, in terms of an ordering, of what a fuzzy, gradual property can be; and second, and non negligible at all, Gödel logic has been proved to be fully compatible with an already proposed and suitable extension of necessity measures for fuzzy events, in the sense that Gödel logic will allow to define a well behaved and featured possibilistic semantics on top of it (see Sections 3 and 4).

Our second objective is to extend the possibilistic logic programming language with a partial matching mechanism between fuzzy propositional variables based on a necessity-like measure. As we have recently proved in [Alsinet and Godo, 2000b], this extension preserves completeness for a particular class of formulas. In our opinion, this is a key feature that justifies by itself the interest of such a logic programming system.

The paper is organized as follows. In Section 2, we present the syntax and the many-valued semantics of the language of propositional Gödel fuzzy logic. In Section 3 we extend the language to allow possibilistic reasoning. In Section 4 we describe the uncertainty sublogic that our proof method can deal with and we prove that it is complete for determining the maximum degree of possibilistic entailment of fuzzy propositional variables. In Section 5, we show how the proof method can be extended to allow a semantical matching between fuzzy propositional variables. And, finally, in the last section we discuss some related work. Because of the lengthy proofs, proofs of propositions can be found in [Alsinet and Godo, 2000a].

## 2   MANY-VALUED SEMANTICS: GÖDEL LOGIC

Following [Hájek, 1998], the language of propositional Gödel fuzzy logic (denoted hereafter G) is built in the usual way from a (countable) set of propositional variables, a conjunction $\wedge$, an implication $\rightarrow$ and the truth constant $\bar{0}$. Further connectives are defined as follows:

$$\varphi \vee \psi \text{ is } ((\varphi \rightarrow \psi) \rightarrow \psi) \wedge ((\psi \rightarrow \varphi) \rightarrow \varphi)$$
$$\neg \varphi \text{ is } \varphi \rightarrow \bar{0}$$
$$\varphi \equiv \psi \text{ is } (\varphi \rightarrow \psi) \wedge (\psi \rightarrow \varphi)$$

The semantics of G is given by *interpretations* $I$ of the propositional variables into the real unit interval $[0, 1]$ which are extended to arbitrary formulas by means of the following rules:

$$I(\bar{0}) = 0$$
$$I(\varphi \wedge \psi) = \min(I(\varphi), I(\psi))$$
$$I(\varphi \rightarrow \psi) = \begin{cases} 1, & \text{if } I(\varphi) \leq I(\psi) \\ I(\psi), & \text{otherwise} \end{cases}$$

For the derived connectives the truth-interpretations take these forms:

$$I(\varphi \vee \psi) = \max(I(\varphi), I(\psi))$$
$$I(\neg \varphi) = \begin{cases} 1, & \text{if } I(\varphi) = 0 \\ 0, & \text{otherwise} \end{cases}$$
$$I(\varphi \equiv \psi) = \begin{cases} 1, & \text{if } I(\varphi) = I(\psi) \\ \min(I(\varphi), I(\psi)), & \text{otherwise} \end{cases}$$

The following is an axiomatization[1] of G:

(A1)  $(\varphi \rightarrow \psi) \rightarrow ((\psi \rightarrow \chi) \rightarrow (\varphi \rightarrow \chi))$
(A2)  $(\varphi \wedge \psi) \rightarrow \varphi$
(A3)  $(\varphi \wedge \psi) \rightarrow (\psi \wedge \varphi)$
(A4)  $(\varphi \wedge (\varphi \rightarrow \psi)) \rightarrow (\psi \wedge (\psi \rightarrow \varphi))$
(A5)  $(\varphi \rightarrow (\psi \rightarrow \chi)) \equiv ((\varphi \wedge \psi) \rightarrow \chi)$
(A6)  $((\varphi \rightarrow \psi) \rightarrow \chi) \rightarrow (((\psi \rightarrow \varphi) \rightarrow \chi) \rightarrow \chi)$
(A7)  $\bar{0} \rightarrow \varphi$
(A8)  $\varphi \rightarrow \varphi \wedge \varphi$

The *deduction rule* of G is modus ponens. The notion of proof is as usual. *Completeness* for G reads as follows: $\varphi$ is provable in G, written $\vdash_G \varphi$, iff $I(\varphi) = 1$ for any interpretation $I$. Furthermore, G enjoys *strong completeness* as well. Namely, let $T$ be an arbitrary theory over G, i.e. just a set of formulas. An interpretation $I$ is a model of $T$ iff $I(\psi) = 1$ for all $\psi \in T$. Then, $T$ proves $\varphi$, written $T \vdash \varphi$, iff $I(\varphi) = 1$ for any interpretation $I$ which is a model of $T$.

## 3   GENERAL POSSIBILISTIC REASONING OVER GÖDEL LOGIC

We have seen that fuzzy propositional variables are suitable for representing imprecise information as in the statement "Peter is *about_35* years old". Now, we are interested in extending the fuzzy propositional language to allow fuzzy reasoning under uncertainty which leads us to a more expressive language. For instance, the statement "it is almost sure that Peter is *about_35* years old" is represented in this setting by a *certainty-weighted fuzzy proposition*

$$(Peter\_is\_about_{35}, 0.9),$$

where the certainty value 0.9 expresses how much is believed the fuzzy statement "Peter is *about_35* years old" in terms of necessity measures.

In general, certainty weights are employed to model statements of the form "$\varphi$ is $\alpha$-certain", where $\varphi$ represents vague, incomplete or imprecise knowledge about the real world. In this framework, this is formalized as "$\varphi$ is certain with a necessity of at least $\alpha$" and is represented through a *certainty-weighted* Gödel logic formula $(\varphi, \alpha)$.

---

[1] Actually, logic G is equivalent to the extension of Intuitionistic logic with the pre-linearity axiom $(\varphi \rightarrow \psi) \vee (\psi \rightarrow \varphi)$.

Within the possibilistic model of uncertainty, belief states are modeled by normalized possibility distributions $\pi : \mathcal{I} \to [0,1]$ on a set of interpretations $\mathcal{I}$. In the standard possibilistic logic, the uncertainty about whether a (Boolean) formula holds true or not is estimated by the necessity measure

$$N([\varphi] \mid \pi) = \inf\{1 - \pi(I) \mid I \in \mathcal{I}, \; I(\varphi) = 0\}.$$

However, in our framework and according to the previous subsection, the truth evaluation of a formula $\varphi$ in each interpretation $I$ is a value $I(\varphi) \in [0,1]$. Therefore, each formula does not induce a crisp set of interpretations, but a fuzzy set of interpretations $[\varphi]$, defining $\mu_{[\varphi]}(I) = I(\varphi)$, for each interpretation $I$. Hence, to measure the uncertainty induced on a formula by a possibility distribution on the set of interpretations $\mathcal{I}$ we have to consider some extension of the notion of necessity measure for fuzzy sets, in particular for fuzzy sets of interpretations. In [Dubois and Prade, 1991] the authors propose to define

$$N([\varphi] \mid \pi) = \inf_{I \in \mathcal{I}} \pi(I) \Rightarrow \mu_{[\varphi]}(I),$$

where $\mu_{[\varphi]}(I) = I(\varphi) \in [0,1]$ and $\Rightarrow$ is the reciprocal of Gödel's many-valued implication, which is defined as $x \Rightarrow y = 1$ if $x \leq y$ and $x \Rightarrow y = 1 - x$, otherwise.

The following remarks are worth noticing. The first one is that this definition is indeed an extension of the classical definition, in the sense that we recover it whenever $[\varphi]$ is a crisp set. The second one is that with this definition, the condition $N([\varphi] \mid \pi) \geq \alpha$ is equivalent to

$$\pi(I) \leq \max(1 - \alpha, \mu_{[\varphi]}(I))$$

for every $I \in \mathcal{I}$, analogously to the crisp case as well. And the third one is that the equivalence

$$N([\varphi] \mid \pi) = 1 \; \text{ iff } \; \pi \leq \mu_{[\varphi]}$$

is also an interesting consequence. It is not difficult to see that this kind of necessity measures on fuzzy sets is characterized by the following set of axioms:

N1  $N(\Omega) = 1$
N2  $N(\emptyset) = 0$
N3  $N(A \cap B) = \min(N(A), N(B))$
   ( $N(\cap_{i \in I} A_i) = \inf_{i \in I} N(A_i)$ )
N4  if $A$ is crisp,
   then $N(A \cup \alpha) = \begin{cases} 1, & \text{if } 1 - \alpha \leq N(A) \\ N(A), & \text{otherwise} \end{cases}$

Now let us go into formal definitions.

**Definition 1 (Possibilistic model)** *Let $\mathcal{I}$ be the set of (many-valued) Gödel interpretations over a given language. A possibilistic model is a normalized possibility distribution $\pi : \mathcal{I} \to [0,1]$ on the set of interpretations $\mathcal{I}$.*

**Definition 2 (Possibilistic satisfaction and entailment)** *A possibilistic model $\pi : \mathcal{I} \to [0,1]$ satisfies a weighted formula $(\varphi, \alpha)$, written $\pi \models (\varphi, \alpha)$, iff $N([\varphi] \mid \pi) \geq \alpha$. Now let $\Gamma$ be a set of weighted formulas. We say that $\Gamma$ entails a weighted formula $(\varphi, \alpha)$, written $\Gamma \models (\varphi, \alpha)$, iff every possibilistic model satisfying all the weighted formulas in $\Gamma$ also satisfies $(\varphi, \alpha)$.*

We propose now a Hilbert-style axiomatization[2] of this logic: axioms of PGL (for Possibilistic Gödel logic) are Gödel logic axioms weighted by 1 plus the triviality axiom $(\varphi, 0)$, and PGL inference rules are :

**Generalized modus ponens:**

$$\frac{(\varphi \to \psi, \alpha) \quad (\varphi, \beta)}{(\psi, \min(\alpha, \beta))}$$

**Weakening:**

$$\frac{(\varphi, \alpha)}{(\varphi, \beta)} \; [\text{if } \beta < \alpha]$$

Then the notion of proof in PGL is as usual and it will be denoted as $\vdash_{PG}$. The soundness of this axiomatic system is given in the next theorem.

**Theorem 1** *PGL is sound with respect the possibilistic entailment, i.e. if $\Gamma \vdash_{PG} (\varphi, \alpha)$ then $\Gamma \models (\varphi, \alpha)$.*

*Proof:* Soundness of the axioms and the weakening rule is straightforward, thus we only prove the soundness of the modus ponens rule. This reduces to check, for all possibilistic distribution $\pi$ on $\mathcal{I}$, that if $N([\varphi \to \psi] \mid \pi) \geq \alpha$ and $N([\varphi] \mid \pi) \geq \beta$, then $N([\psi] \mid \pi) \geq \min(\alpha, \beta)$. The two conditions amount to, for each interpretation $I \in \mathcal{I}$, $\pi(I) \Rightarrow I(\varphi \to \psi) \geq \alpha$ and $\pi(I) \Rightarrow I(\varphi) \geq \beta$. Thus, $\pi(I) \Rightarrow \min(I(\varphi \to \psi), I(\varphi)) \geq \min(\alpha, \beta)$. But, by Gödel semantics, $\min(I(\varphi \to \psi), I(\varphi)) = \min(I(\psi), I(\varphi)) \leq I(\psi)$. Therefore, we have that $\pi(I) \Rightarrow I(\psi) \geq \min(\alpha, \beta)$ for each interpretation $I \in \mathcal{I}$, and thus, $N([\psi] \mid \pi) \geq \min(\alpha, \beta)$. □

Unfortunately we have not been able so far to prove whether PGL is complete or incomplete. So it remains as an open question to be solved in the near future.

## 4 A POSSIBILISTIC LOGIC PROGRAMMING LANGUAGE

In the previous section we have defined PGL, a general possibilistic logic over the many-valued Gödel logic. Our aim in this section is, as in classical propositional logic programming systems, to define a sublanguage for logic programming which would enable us to design an efficient proof algorithm, based on a complete calculus for computing the maximum degree of possibilistic entailment of a propositional variable, called *goal*, from a set of weighted formulas.

---

[2]Notice the analogy with the classical Possibilistic logic [Dubois et al., 1994c], here we have just replaced the axioms of classical propositional logic with those of many-valued Gödel logic.

To this end, we restrict ourselves to a Horn-rule sublanguage of the logic G, i.e. to formulas of the form:

$$p_1 \wedge \cdots \wedge p_k \rightarrow q$$

with $k \geq 0$, where $p_1, \ldots, p_k, q$ are propositional variables, in the traditional logic programming style. As usual, we shall refer to the conclusion $q$ and the set of premises $p_1, \ldots, p_k$ as the *head* and the *body*, respectively. We distinguish between two types of formulas in this sublanguage: **fact** when $k = 0$ (empty body) and are simply written $q$, and **rule**, otherwise.

**Definition 3 (PGL clause)** *A PGL clause is a pair of the form $(\varphi, \alpha)$, where $\varphi$ is either a fact or a rule and $\alpha \in [0, 1]$ is a lower bound on the belief on $\varphi$ in terms of necessity measures.*

For PGL clauses we shall develop a simple and efficient calculus which will not need the whole logical apparatus of PGL of the previous section. But before we need to introduce some extra definitions and results.

**Definition 4 (Maximum degree of possibilistic entailment)** *The maximum degree of possibilistic entailment of a goal $q$ from a set of PGL clauses $P$, denoted by $\|q\|_P$, is the greatest lower bound $\alpha \in [0, 1]$ on the belief on $q$ such that $P \models (q, \alpha)$. Thus, $\|q\|_P = \sup\{\alpha \in [0, 1] \mid P \models (q, \alpha)\}$.*

**Theorem 2** *The maximum degree of possibilistic entailment of a goal $q$ from a set of PGL clauses $P$ is the least necessity evaluation of $q$ given by the models of $P$. Thus, $\|q\|_P = \inf\{N([q] \mid \pi) \mid \pi \models P\}$.*

*Proof:* We define $\alpha_1 = \sup\{\alpha \in [0, 1] \mid P \models (q, \alpha)\}$ and $\alpha_2 = \inf\{N([q] \mid \pi) \mid \pi \models P\}$.

- $\alpha_2 \geq \alpha_1$: As $\alpha_1 = \sup\{\alpha \in [0, 1] \mid P \models (q, \alpha)\}$ we have that $P \models (q, \alpha)$, for all $\alpha < \alpha_1$. Then, for every model $\pi$ of $P$ we have that $N([q] \mid \pi) \geq \alpha$ for all $\alpha < \alpha_1$. Thus, $\alpha_2 = \inf\{N([q] \mid \pi) \mid \pi \models P\} \geq \alpha$, for all $\alpha < \alpha_1$, hence, $\alpha_2 \geq \alpha_1$.

- $\alpha_2 \leq \alpha_1$: As $\alpha_2 = \inf\{N([q] \mid \pi) \mid \pi \models P\}$ we have that $N([q] \mid \pi) \geq \alpha_2$ for all model $\pi$ of $P$, that is $P \models (q, \alpha_2)$, and thus, $\alpha_2 \leq \sup\{\alpha \in [0, 1] \mid P \models (q, \alpha)\} = \alpha_1$. □

**Corollary 1** *Let $P$ be a of PGL clauses and let $s$ be a propositional variable. Then, $P \models (s, \|s\|_P)$.*

To provide our possibilistic logic programming language with a complete calculus for determining the maximum degree of possibilistic entailment we only need the triviality axiom and a particular instance of the generalized modus ponens rule introduced in the previous section:

**Axiom:** $(\varphi, 0)$

**Generalized modus ponens:**

$$\frac{(p_1 \wedge \cdots \wedge p_k \rightarrow q, \gamma)}{(q, \min(\gamma, \beta_1, \ldots, \beta_k))}$$

Obviously, the axiom is a valid PGL clause and the inference rule is sound as already proved in the previous section.

**Definition 5 (Degree of deduction)** *A goal $q$ is deduced with a degree of deduction $\alpha$ from a set of PGL clauses $P$, denoted by $P \vdash^* (q, \alpha)$, iff there exists a finite sequence of PGL clauses $C_1, \ldots, C_m$ such that $C_m = (q, \alpha)$ and, for each $i \in \{1, \ldots, m\}$, it holds that $C_i \in P$, $C_i$ is an instance of the axiom or $C_i$ is obtained by applying the above inference rule to previous clauses in the sequence.*

Next, we define the syntactic counterpart of maximum degree of possibilistic entailment.

**Definition 6 (Maximum degree of deduction)** *The maximum degree of deduction of a goal $q$ from a set of PGL clauses $P$, denoted $|q|_P$, is the greatest $\alpha \in [0, 1]$ such that $P \vdash^* (q, \alpha)$.*

As the only inference rule of our proof method is the generalized modus ponens, within the framework of logic programming in which $P$ is always a finite set of PGL clauses, there exists a finite number of proofs of a goal $q$ from $P$, and thus, the above definition turns into $|q|_P = \max\{\alpha \in [0, 1] \mid P \vdash^* (q, \alpha)\}$.

The following propositions are needed to prove completeness.

**Proposition 1** *If the only formulas in $P$ with head $q$ are recursive[3], then $\|q\|_P = |q|_P = 0$.*

**Proposition 2** *Let $(p, \beta)$ and $(p \rightarrow q, \gamma)$ be two PGL clauses such that $p \neq q$. Then, $\|q\|_{\{(p,\beta),(p \rightarrow q,\gamma)\}} = \min(\beta, \gamma)$.*

**Proposition 3** *Let $P$ be a set of PGL clauses and let $(r \rightarrow q, \gamma)$ be a clause of $P$. If $\|q\|_P > \|q\|_{P \setminus \{(r \rightarrow q,\gamma)\}}$, then $\|q\|_P = \|q\|_{\{(r,\alpha_r),(r \rightarrow q,\gamma)\}}$, where $\alpha_r = \|r\|_P$.*

**Proposition 4** *Let $P$ be a set of PGL clauses and let $(q \rightarrow p, \gamma)$ be a clause of $P$. It holds that $\|q\|_P = \|q\|_{P \setminus \{(q \rightarrow p,\gamma)\}}$.*

**Theorem 3 (Completeness)** *Let $P$ be a set of PGL clauses and let $q$ be a goal. Then, $\|q\|_P = |q|_P$.*

*Proof:* By the soundness of the modus ponens inference rule, we have $\|q\|_P \geq |q|_P$. Therefore, we must prove $\|q\|_P \leq |q|_P$ and we proceed by induction on $n$, where $n$ is the number of clauses of $P$.

---
[3] A recursive formula is of the form $p_1 \wedge \cdots \wedge p_k \wedge q \rightarrow q$ with $k \geq 0$.

If $n = 1$, then it must be that $P$ contains only either one certainty-weighted fact or rule. We assume that $q$ occurs in $P$ as the head of a non recursive formula; otherwise, by Proposition 1, we have $\|q\|_P = |q|_P = 0$.

Suppose that $P$ contains only the certainty-weighted fact $(q, \gamma)$. Let $I_0$ and $I_1$ be two interpretations such that $I_0(q) < 1 - \gamma$ and $I_1(q) = 1$. Now, let $\pi$ be a possibility distribution with the following definition:

$$\pi(I) = \begin{cases} 1, & \text{if } I = I_1 \\ 1 - \gamma, & \text{if } I = I_0 \\ 0, & \text{otherwise} \end{cases}$$

It is easy to check that $\pi \models (q, \gamma)$ and $N([q] \mid \pi) = \gamma$. Then, by Theorem 2, we have $\|q\|_P = \inf\{N([q] \mid \pi) \mid \pi \models (q, \gamma)\} \leq \gamma$. Now, $|q|_P = \max\{\alpha \in [0,1] \mid (q, \gamma) \vdash^* (q, \alpha)\} = \gamma$, and thus, $\|q\|_P \leq |q|_P$.

Suppose that $P$ contains only the certainty-weighted rule $(r \to q, \gamma)$. Let $I_0$ be an interpretation such that $I_0(r) \leq I_0(q) < 1$ and let $\pi$ be a possibility distribution with the following definition:

$$\pi(I) = \begin{cases} 1, & \text{if } I = I_0 \\ 0, & \text{otherwise} \end{cases}$$

It is easy to check that $\pi \models (r \to q, \gamma)$ and $N([q] \mid \pi) = 0$. Then, by Theorem 2, we have $\|q\|_P = \inf\{N([q] \mid \pi) \mid \pi \models (r \to q, \gamma)\} = 0$, and thus, $\|q\|_P \leq |q|_P$.

Suppose now that for any set $P'$ that contains $n$ clauses it holds that $\|s\|_{P'} \leq |s|_{P'}$ for any propositional variable $s$, and suppose that $P$ contains $n + 1$ clauses. Since we are assuming that $q$ occurs in $P$ as the head of a non recursive formula, let $(\varphi, \gamma)$ be a clause of $P$ such that $\varphi$ is a non recursive formula with head $q$ and $\gamma \geq \delta$, for any clause $(\psi, \delta)$ of $P$ such that $\psi$ is a non recursive formula with head $q$. We distinguish two cases:

**Case** $(\varphi, \gamma) = (q, \gamma)$. Let $I_0$ and $I_1$ be two interpretations such that $I_0(q) < 1 - \gamma$ and $I_0(p) = 1$ for any propositional variable $p \neq q$, and $I_1(s) = 1$ for any propositional variable $s$. Now, let $\pi$ be a possibility distribution with the following definition:

$$\pi(I) = \begin{cases} 1, & \text{if } I = I_1 \\ 1 - \gamma, & \text{if } I = I_0 \\ 0, & \text{otherwise} \end{cases}$$

Since $\gamma \geq \delta$ for any non recursive clause $(\psi, \delta)$ of $P$ such that $q$ is the head of $\psi$, we have that $\pi \models P$ and $N([q] \mid \pi) = \gamma$. Therefore, by Theorem 2, we have $\|q\|_P = \inf\{N([q] \mid \pi) \mid \pi \models P\} \leq \gamma$. Now, $|q|_P = \max\{\alpha \in [0,1] \mid P \vdash^* (q, \alpha)\} \geq \gamma$, and thus, $\|q\|_P \leq |q|_P$.

**Case** $(\varphi, \gamma) = (p \to q, \gamma)$. If we define $P' = P \setminus \{(p \to q, \gamma)\}$ we have that $\|q\|_P \geq \|q\|_{P'}$.

If $\|q\|_P = \|q\|_{P'}$, by the induction hypothesis, we have that $\|q\|_{P'} \leq |q|_{P'}$, and thus, $\|q\|_P \leq |q|_P$.

If $\|q\|_P > \|q\|_{P'}$, by Proposition 3, $\|q\|_P = \|q\|_{\{(p, \alpha_p), (p \to q, \gamma)\}}$, where $\alpha_p = \|p\|_P$, and, by Proposition 2, $\|q\|_{\{(p, \alpha_p), (p \to q, \gamma)\}} = \min(\alpha_p, \gamma)$. Now, by Proposition 4, $\|p\|_P = \|p\|_{P'}$ and, by the induction hypothesis, $\|p\|_{P'} \leq |p|_{P'}$. Since $|p|_{P'} \leq |p|_P$ and $(p \to q, \gamma) \in P$, applying the modus ponens inference rule, we get $P \vdash^* (q, \min(|p|_P, \gamma))$ with $\min(|p|_P, \gamma) \geq \min(\alpha_p, \gamma)$. Then, $|q|_P = \max\{\alpha \in [0,1] \mid P \vdash^* (q, \alpha)\} \geq \min(|p|_P, \gamma)$, and thus, $|q|_P \geq \min(\alpha_p, \gamma) = \|q\|_P$.  □

In the particular case that we do not allow recursive formulas in the language, the underlying uncertainty logic of our logic programming system is syntactical equivalent to the family of infinitely-valued propositional logics the interpreter defined in [Escalada-Imaz and Manyà, 1995] can deal with. The interpreter is based on a backward proof algorithm for computing the maximum degree of deduction of a propositional variable from a set of formulas whose worst-case time complexity is linear in the total number of occurrences of propositional variables in the set of formulas. We show bellow an example of PGL clauses the interpreter can deal with.

**Example 1** *The maximum degree of deduction of the goal $friend\_Mary\_John$ from the set of clauses*

$P=\{$ $(Mary\_is\_young, 0.8),$
$(John\_is\_young, 0.9),$
$(Mary\_is\_young \wedge John\_is\_young \to$
$friend\_Mary\_John, 0.6)$ $\},$

*is 0.6 which corresponds with the deduction degree computed by the interpreter when taking as triangular norm the min-conjunction function and as implication Gödel's many-valued implication function.*

## 5 ADDING FUZZY UNIFICATION

Our aim is to extend the calculus of our possibilistic language to allow a semantical matching between fuzzy propositional variables based on a necessity evaluation of fuzzy events. For instance, given the set of PGL clauses

$P=\{$ $(Mary\_is\_young, 0.8),$
$(John\_is\_about_{16}, 0.9),$
$(Mary\_is\_young \wedge John\_is\_young \to$
$friend\_Mary\_John, 0.6)$ $\},$

where $John\_is\_young$ and $John\_is\_about_{16}$ are two fuzzy propositional variables, the maximum degree of deduction of the goal $friend\_Mary\_John$ from $P$ is 0 unless we be able to compute the necessity evaluation of the propositional variable $John\_is\_young$ from the fact that the propositional variable $John\_is\_about_{16}$ is certain with a necessity of at lest 0.9.

To tackle the fuzzy unification problem within our possibilistic framework we are lead (i) to fix a domain and an interpretation of fuzzy propositional variables in terms of its membership functions, otherwise we would not be able to identify when two propositional variables $A$ and $B$ are related and, furthermore, we would not be able to reason about the certainty of the propositional variable $B$ from the fact that the domain-related propositional variable $A$ is certain with a necessity of at least $\alpha$; and (ii) to define some measure to compute the necessity evaluation of a fuzzy propositional variable $B$ based on a domain-related fuzzy propositional variable $A$.

## 5.1 PROPOSITIONAL VARIABLES WITH FINITE DOMAINS

With respect the underlying uncertainty logic described in Section 2, the main difference is that now we attach propositional variables with a *sort*. In doing so, we are introducing a minor change in the semantics. Many-valued interpretations should map a sort into a non-empty domain and a propositional variable into a value of its domain, and thus, in turn we need to provide a new notion of interpretation.

A more rigorous approach should be to define a first-order language with typed regular predicates and sorted fuzzy constants (cf. [Alsinet et al., 1999]) which would allow us to represent, for instance, the fuzzy statement "Mary is young" as $age(Mary, young)$. However, since variables and function symbols are not allowed in the language, fuzzy propositional variables give us a more simple representation model without lost of expressiveness.

**Definition 7 (Extended many-valued interpretation)** *An interpretation $I = (U, i, m)$ maps:*

1. *each sort $\sigma$ into a non-empty domain $U_\sigma$;*

2. *all propositional variables $p$ of sort $\sigma$ into a same value $i(p) \in U_\sigma$; and*

3. *a propositional variable $p$ of sort $\sigma$ into a (normalized) fuzzy set $m(p) : U_\sigma \to [0, 1]$.*

Notice that an interpretation $I = (U, i, m)$ is disjunctive in the sense that $i(p)$ is a unique value of the domain $U_\sigma$ for any propositional variable $p$ of sort $\sigma$, and $i(p) \neq i(p')$ iff $p$ and $p'$ are of different sorts.

In what follows, we shall denote by $\mu_{m(p)}$ the membership function of $m(p)$. The *truth value* of a propositional variable $p$ under an interpretation $I = (U, i, m)$, denoted by $I(p)$, is computed as

$$I(p) = \mu_{m(p)}(i(p)).$$

These truth evaluations extend to rules in the usual way by using the truth functions described in Section 2.

Remark that the truth value of a propositional variable $p$ under an interpretation $I = (U, i, m)$ depends not only on the value $i(p)$ assigned to $p$, but on the fuzzy set $m(p)$. Therefore, in order to measure the certainty of a sorted propositional variable in a possibilistic model we cannot take into account any possible interpretation, but only those which share a common interpretation of propositional variables, and hence which also share their domain. This leads us to define the notion of context (cf. [Alsinet et al., 1999])

**Definition 8 (Context)** *Let $U$ be a collection of non-empty domains and let $m$ be an interpretation of propositional variables over $U$ such that for every propositional variable $p$ there exist $u, v \in U_\sigma$, $\sigma$ being the sort of $p$, such that $\mu_{m(p)}(u) = 0$ and $\mu_{m(p)}(v) = 1$. We define the context determined by $U$ and $m$, denoted by $\mathcal{I}_{U,m}$, as the set of extended many-valued interpretations having $U$ as domain and $m$ as interpretation of propositional variables. Thus,*

$$\mathcal{I}_{U,m} = \{I \in \mathcal{I} \mid I = (U, i, m)\},$$

*where $\mathcal{I}$ denotes the set of all extended many-valued interpretations.*

Remark that given a context $\mathcal{I}_{U,m}$ and a propositional variable $p$, there exist at least two interpretations $I_0, I_1 \in \mathcal{I}_{U,m}$ such that $I_0(p) = 0$ and $I_1(p) = 1$.

Let us briefly discuss the reason for defining the notion of context by means of an example.

**Example 2** *Let $age\_Mary\_around_{19}$ be a propositional variable of sort $Mary\_years\_old$ and let $I_0 = (U, i, m_0)$ and $I_1 = (U, i, m_1)$ be two interpretations such that*

- $U_{Mary\_years\_old} = [0, 120](years)$,

- $i(age\_Mary\_around_{19}) = 20$,

- $m_0(age\_Mary\_around_{19}) = [17; 18; 20; 21]$ *-trapezoidal fuzzy set[4]– and*

- $m_1(age\_Mary\_around_{19}) = [18; 19; 19; 20]$.

*Although the age of Mary is the same value in both interpretations, we have a different truth value in each interpretation depending on the membership function of the fuzzy set assigned to the propositional variable. Thus,*

$$I_0(age\_Mary\_around_{19}) = \mu_{[17;18;20;21]}(20) = 1$$

*and*

$$I_1(age\_Mary\_around_{19}) = \mu_{[18;19;19;20]}(20) = 0.$$

*Then, for instance, the possibility distributions $\pi$ that satisfy the PGL clause*

$$(age\_Mary\_around_{19}, 1)$$

*are those such that $\pi(I_0) \leq 1$ and $\pi(I_1) = 0$, and thus, $I_0$ can be fully plausible while $I_1$ is inadmissible, however, Mary is twenty years old in both interpretations.*

---

[4]We represent a trapezoidal fuzzy set as $[t_1; t_2; t_3; t_4]$, where the interval $[t_1, t_4]$ is the support and the interval $[t_2, t_3]$ is the core.

Therefore, when fixing a particular context we are ensuring that belief states modeled by normalized possibility distributions on a set of possible interpretations (or possible states) are consistent, in the sense that possible states are sharing a common view of the real world.

Finally, the notion of possibilistic satisfaction can be easily extended in a particular context $\mathcal{I}_{U,m}$ in the following way. Given a context $\mathcal{I}_{U,m}$, the models of a set of PGL clauses $P$ are the normalized possibility distributions on the set of extended many-valued interpretations $\mathcal{I}_{U,m}$ that satisfy all the clauses in $P$.

## 5.2 EXTENDED INFERENCE WITH POSSIBILISTIC PATTERN MATCHING

To provide PGL with a semantical matching mechanism we need a measure for computing the necessity evaluation of the propositional variable $B$ based on the domain-related propositional variable $A$. Furthermore, this measure should enable us to include new inference patterns in our previous calculus of Section 4 in order to keep completeness for determining the maximum degree of possibilistic entailment of a goal form a set of PGL clauses in a particular context.

Again there are several alternatives. After a careful analysis we have chosen the same type of measure used when defining the possibilistic semantics for PGL. Namely, given a context $\mathcal{I}_{U,m}$, the necessity evaluation of $B$ based on the domain-related propositional variable $A$ is defined as

$$N(m(B) \mid m(A)) = \inf_{u \in U_\sigma} \mu_{m(A)}(u) \Rightarrow \mu_{m(B)}(u),$$

where $\Rightarrow$ is the reciprocal of Gödel's many-valued implication.

At this point we are ready to extend the calculus to allow a semantical matching of propositional variables with finite domains through a possibilistic pattern matching measure based on a necessity evaluation of fuzzy events. Although there may exist several approaches, we have finally decided to extend the calculus with three inference rules. Given a context $\mathcal{I}_{U,m}$, these inference rules are defined as follows:

**Semantical unification:**
$$\frac{(p, \alpha)}{(p', \min(\alpha, \beta))} \; [SU],$$
where $\beta = N(m(p') \mid m(p))$.

**Intersection:**
$$\frac{(p_1, \alpha), (p_2, \beta)}{(p', \min(\alpha, \beta))} \; [IN],$$
if $\mu_{m(p')} \geq \min(\mu_{m(p_1)}, \mu_{m(p_2)})$.

**Resolving Uncertainty:**
$$\frac{(p, \alpha)}{(p', 1)} \; [UN],$$
if $\mu_{m(p')} \geq \max(1 - \alpha, \mu_{m(p)})$.

**Theorem 4** *The SU, IN and UN inference rules are sound in a context $\mathcal{I}_{U,m}$ with respect to the possibilistic entailment of PGL clauses.*

*Proof:* Soundness of the IN and UN rules is straightforward, thus we only prove the soundness of the SU rule. We must prove that for all possibilistic distribution $\pi : \mathcal{I}_{U,m} \to [0,1]$, if $\pi \models (p, \alpha)$ then $\pi \models (p', \min(\alpha, \beta))$, where $\beta = N(m(p') \mid m(p))$. Assume that $\pi \models (p, \alpha)$. This means that, $N([p] \mid \pi) \geq \alpha$, and thus, $\pi(I) \Rightarrow I(p) \geq \alpha$, for all interpretation $I \in \mathcal{I}_{U,m}$. Then, we have the following consecutive inequalities:

1. $\pi(I) \Rightarrow \mu_{m(p)}(i(p)) \geq \alpha$

2. $\min(\pi(I) \Rightarrow \mu_{m(p)}(i(p)),$
   $\mu_{m(p)}(i(p)) \Rightarrow \mu_{m(p')}(i(p'))) \geq$
   $\min(\alpha, \mu_{m(p)}(i(p)) \Rightarrow \mu_{m(p')}(i(p')))$

3. $\min(\pi(I) \Rightarrow \mu_{m(p)}(i(p)),$
   $\mu_{m(p)}(i(p)) \Rightarrow \mu_{m(p')}(i(p'))) \leq$
   $\pi(I) \Rightarrow \mu_{m(p')}(i(p'))$

4. $\pi(I) \Rightarrow \mu_{m(p')}(i(p')) \geq$
   $\min(\alpha, \mu_{m(p)}(i(p)) \Rightarrow \mu_{m(p')}(i(p')))$

5. $\inf_{I \in \mathcal{I}_{U,m}} \pi(I) \Rightarrow \mu_{m(p')}(i(p')) \geq$
   $\inf_{I \in \mathcal{I}_{U,m}} \min(\alpha, \mu_{m(p)}(i(p)) \Rightarrow \mu_{m(p')}(i(p')))$

6. $\inf_{I \in \mathcal{I}_{U,m}} \pi(I) \Rightarrow \mu_{m(p')}(i(p')) \geq$
   $\min(\alpha, \inf_{I \in \mathcal{I}_{U,m}} \mu_{m(p)}(i(p)) \Rightarrow \mu_{m(p')}(i(p')))$

7. $N([p'] \mid \pi) \geq$
   $\min(\alpha, \inf_{u \in U_\sigma} \mu_{m(p)}(u) \Rightarrow \mu_{m(p')}(u))$

8. $N([p'] \mid \pi) \geq \min(\alpha, N(m(p') \mid m(p)))$

Thus, $\pi \models (p', \min(\alpha, N(m(p') \mid m(p))))$ as well. □

Moreover, for a given a context $\mathcal{I}_{U,m}$, the modus-ponens style calculus extended with the SU, IN and UN inference rules is *complete* for determining the maximum degree of possibilistic entailment of a goal from a particular class of PGL clauses called well-formed and satisfiable PGL programs. The formalization of the notion of well-formed and satisfiable PGL program and the proof of completeness for this class of PGL clauses can be found in [Alsinet and Godo, 2000b].

**Example 3** *Let $John\_is\_about_{16}$, $John\_is\_14\_16$ and $John\_is\_16\_18$ be three propositional variables of sort $John\_years\_old$, and let $\mathcal{I}_{U,m}$ be a context such that*

- $U_{John\_years\_old} = [0, 120](years)$,

- $m(John\_is\_about_{16}) = [14; 16; 16; 18]$,

- $m(John\_is\_14\_16) = [12; 14; 16; 18]$ and

- $m(John\_is\_16\_18) = [14; 16; 18; 20]$.

*Now, given the set of PGL clauses*

$$P=\{ (John\_is\_14\_16, 1),\\ (John\_is\_16\_18, 1) \},$$

*we have that*

$$N(m(John\_is\_about_{16}) \mid m(John\_is\_14\_16)) = 0$$

*and*

$$N(m(John\_is\_about_{16}) \mid m(John\_is\_16\_18)) = 0.$$

*However, each model $\pi : \mathcal{I}_{U,m} \to [0,1]$ of $P$ verifies that*

$$\pi(I) \leq \min(I(John\_is\_14\_16), I(John\_is\_16\_18)),$$

*for all $I \in \mathcal{I}_{U,m}$. Therefore, because of the three propositional variables have the same domain, we have that $i(John\_is\_14\_16) = i(John\_is\_16\_18) = i(John\_is\_about_{16})$, for all $I \in \mathcal{I}_{U,m}$. Then, because of this particular interpretation $m$ of propositional variables,*

$$\pi(I) \leq \mu_{m(John\_is\_about_{16})}(i(John\_is\_about_{16})),$$

*for all $I \in \mathcal{I}_{U,m}$. Hence, $P \models (John\_is\_about_{16}, 1)$, and thus, $\|John\_is\_about_{16}\|_P = 1$. On the other hand, applying the IN rule to $(John\_is\_14\_16, 1)$ and $(John\_is\_16\_18, 1)$ we deduce $(John\_is\_about_{16}, 1)$, and thus, $|John\_is\_about_{16}|_P = \|John\_is\_about_{16}\|_P$.*

## 6 RELATED WORK

The introduction of fuzzy constants in logic programming languages was suggested in the early eighties by [Cayrol et al., 1982] and [Bel et al., 1986] with the aim of including fuzzy values in a pattern matching procedure. Subsequently, [Umano, 1987] defined a fuzzy pattern matching process using the extension principle for one and two variate functions, and in [Baldwin et al., 1995] the authors implemented a semantic unification procedure based on the theory of mass assignments which allows a unified framework for the treatment of fuzzy and probabilistic data. [Godo and Vila, 1995] proposed a possibilistic-based logic to deal with fuzzy temporal constraints based on many-valued semantics and a necessity-like measure to allow a pattern matching mechanism between fuzzy temporal constraints. [Virtanen, 1998] defined a fuzzy unification algorithm based on fuzzy equality relations and [Arcelli et al., 1998] proposed three different kinds of unification in the fuzzy context: the first one is based on similarity relations, the second one identifies similar objects through an equivalence relation and the last one uses "semantic constraints" for defining a more flexible unification. More recently, [Gerla and Sessa, 1999] formalized a methodology for transforming an interpreter for SLD Resolution into an interpreter that computes on abstract values which express similarity properties on the set of predicate and function symbols of a classical first-order language and [Formato et al., 2000] extended the unification algorithm of Martelli-Montanari to allow a partial matching between crisp constants through similarity relations. Finally, [Dubois et al., 1998] proposed an extension of possibilistic logic dealing with fuzzy constants and fuzzily restricted quantifiers (called PLFC), and [Alsinet et al., 1999] provided PLFC logic with a formal semantics and a sound resolution-style calculus by refutation.

The main differences between this framework and PLFC [Alsinet et al., 1999] are (i) at the level of the syntax and semantics of the language, (ii) at the level of providing the language with a complete calculus, and (iii) at the level of defining a semantical unification mechanism between fuzzy propositions.

In PLFC, formulas are pairs of the form $(\varphi(\bar{x}), f(\bar{y}))$, where $\bar{x}$ and $\bar{y}$ denote sets of free and implicitly universally quantified variables and $\bar{y} \supseteq \bar{x}$, $\varphi(\bar{x})$ is a disjunction of literals with fuzzy constants, and $f(\bar{y})$ is a valid valuation function which expresses the certainty of $\varphi(\bar{x})$ in terms of necessity measures. Basically, valuation functions $f(\bar{y})$ are constant values and variable weights which are not considered in our possibilistic language. Variable weights [Dubois et al., 1994a; 1994b] are suitable for modeling statements of the form "the more $x$ is $A$ (or $x$ belongs to $A$), the more certain is $p(x)$", where $A$ is a fuzzy set. This is formalized in PLFC as, for all $x$, "$p(x)$ is true with a necessity of at least $\mu_A(x)$", and is represented as $(p(x), A(x))$. When $A$ is imprecise but not fuzzy, the interpretation of such a formula is just "$\forall x \in A, p(x)$". So $A$ acts as a (flexible if it is fuzzy) restriction on the universal quantifier. On the other hand, fuzzy constants in PLFC can be seen as (flexible) restrictions on an existential quantifier. In general, "$L(B)$ is true at least to degree $\alpha$", is represented in PLFC as $(L(B), \alpha)$, where $L$ is either a positive or a negative literal and $B$ is a fuzzy set. For instance, if $B$ is imprecise but not fuzzy, $(p(B), \alpha)$ and $(\neg p(B), \alpha)$ have to be read as "$\exists x \in B, p(x)$" and "$\exists x \in B, \neg p(x)$", respectively.

Therefore, because of the calculus for PLFC is defined by refutation through a generalized resolution rule between positive and negative literals, the unification (in the classical sense) between fuzzy constants is not allowed. For instance, from

s1: $(\neg p(A) \lor \psi, 1)$ and s2: $(p(A), 1)$,

which, if $A$ is not fuzzy, are interpreted respectively as

"$(\exists x \in A, \neg p(x)) \lor \psi$" and "$\exists x \in A, p(x)$",

we can infer $\psi$ iff $A$ is a precise constant. However, a semantical unification between fuzzy events is performed through variable weights and fuzzy constants. For instance, from

s1: $(\neg p(x) \lor \psi(x), A(x))$ and s2: $(p(B), 1)$,

which, if $A$ and $B$ are not fuzzy, are interpreted respectively as

"$\forall x \in A, \neg p(x) \vee \psi(x)$" and "$\exists x \in B, p(x)$",

we infer $(\psi(B), N(A \mid B))$, where $N(A \mid B)$ is a necessity evaluation of fuzzy events.

In our current setting, fuzzy propositional variables are interpreted in the same way as fuzzy constants in PLFC but, in contrast to PLFC, we do not have negative literals in the language and we have provided the language with a complete modus ponens-style calculus for determining the maximum degree of possibilistic entailment of a propositional variable from a set of formulas which can be extended, through a semantical unification inference rule, to allow a semantical matching between propositional variables.

Concerning the semantics, because of the fuzzy information, the truth evaluation of formulas is many-valued in both languages, and belief states are modeled by normalized possibility distributions on the set of many-valued interpretations, also in both languages. However, the basic connectives of PLFC are negation $\neg$ and disjunction $\vee$ while in our language, they are conjunction $\wedge$ and implication $\rightarrow$, and the semantics for the two sets of connectives are not equivalent, i.e. the two sets of connectives are not inter-definable. Moreover, the extended necessity measure for fuzzy sets suggested by Dubois and Prade in [Dubois and Prade, 1991], which is used in this language for setting the possibilistic semantics of formulas, is different from the one used in PLFC, although both are extensions of the standard necessity measure for crisp sets.

Finally, the proof method for PLFC is based on refutation through a resolution rule, a fusion rule already proposed in [Dubois et al., 1998] and a merging rule. During the proof process, the merging rule must be applied after every resolution step, and thus, the proof algorithm cannot be oriented to a resolvent clause and therefore, the search space consists of all possible orderings of the literals in the input program. In contrast, in our current setting, the proof method is oriented to propositional variables (goals) and can be performed in a bottom-up manner through a generalized modus ponens inference rule. In particular, when the semantical unification between fuzzy propositional variables is emulated by means of the following non-logical axiom

$$(p \rightarrow p', N(m(p') \mid m(p))),$$

we can easily adapt the interpreter proposed in [Escalada-Imaz and Manyà, 1995] for computing $N(m(p') \mid m(p))$ in a particular context. However, as we have seen, this approach is not powerful enough for determining the maximum degree of possibilistic entailment of a goal $q$ from a set of PGL clauses $P$.

## Acknowledgments

We would like to thank the anonymous referees for their comments and suggestions. This research was partially supported by the project SMASH (TIC96-1038-C04-01/03) funded by the CICYT.